\documentclass[10pt,twocolumn,letterpaper]{article}

\usepackage[pagenumbers]{wacv} 

\usepackage{graphicx}
\usepackage{amsmath}
\usepackage{amssymb}
\usepackage{booktabs}
\usepackage{multirow}
\usepackage{makecell}
\usepackage{pifont}
\newcommand{\xmark}{\ding{55}}
\usepackage{arydshln}
\usepackage{tablefootnote}
\usepackage{threeparttable}
\usepackage[linesnumbered,ruled,vlined]{algorithm2e}

\SetCommentSty{mycommfont}
\SetKwInput{KwInput}{Input}                
\SetKwInput{KwOutput}{Output}              
\SetKwInput{KwFunctions}{Functions}         
\SetKwInput{KwTrainables}{Trainable Parameters}
\SetKwInput{KwStopgrad}{Stop Gradient} 
\SetKwInput{KwHyperparameters}{Hyperparameters}    
\SetKwInput{KwSamples}{Samples}

\usepackage{array} 
\newcolumntype{P}[1]{>{\centering\arraybackslash}p{#1}}

\usepackage[dvipsnames]{xcolor}

\usepackage[pagebackref,breaklinks,colorlinks]{hyperref}

\definecolor{NIcustomgreen}{rgb}{0.0, 0.6, 0.0}
\newcommand{\NIgreen}[1]{\textcolor{NIcustomgreen}{#1}}
\definecolor{NIcustomred}{rgb}{0.8, 0.0, 0.0}
\newcommand{\NIred}[1]{\textcolor{NIcustomred}{#1}}
\definecolor{NIcustomblue}{rgb}{0.0, 0.0, 0.8}

\definecolor{NImagenta}{rgb}{1.0, 0.0, 1.0}
\definecolor{NIgray}{gray}{0.5}

\usepackage[capitalize]{cleveref}
\crefname{section}{Sec.}{Secs.}
\Crefname{section}{Section}{Sections}
\Crefname{table}{Table}{Tables}
\crefname{table}{Tab.}{Tabs.}

\def\wacvPaperID{290} 
\def\confName{WACV}
\def\confYear{2025}

\begin{document}




\title{Foundation X: Integrating Classification, Localization, and Segmentation through Lock-Release Pretraining Strategy for Chest X-ray Analysis}


\author{
Nahid Ul Islam$^{1}$ \quad DongAo Ma$^{1}$ \quad Jiaxuan Pang$^{1}$ \quad Shivasakthi Senthil Velan$^{1}$\\
Michael Gotway$^{2}$ \quad Jianming Liang$^{1}$\\
$^{1}$Arizona State University, USA \quad $^{2}$Mayo Clinic, USA\\
}

\maketitle

\begin{abstract}
    Developing robust and versatile deep-learning models is essential for enhancing diagnostic accuracy and guiding clinical interventions in medical imaging, but it requires a large amount of annotated data. The advancement of deep learning has facilitated the creation of numerous medical datasets with diverse expert-level annotations. Aggregating these datasets can maximize data utilization and address the inadequacy of labeled data. However, the heterogeneity of expert-level annotations across tasks such as classification, localization, and segmentation presents a significant challenge for learning from these datasets. To this end, we introduce \textbf{Foundation X}, an end-to-end framework that utilizes diverse expert-level annotations from numerous public datasets to train a foundation model capable of multiple tasks including classification, localization, and segmentation. To address the challenges of annotation and task heterogeneity, we propose a Lock-Release pretraining strategy to enhance the cyclic learning from multiple datasets, combined with the student-teacher learning paradigm, ensuring the model retains general knowledge for all tasks while preventing overfitting to any single task. To demonstrate the effectiveness of Foundation X, we trained a model using 11 chest X-ray datasets, covering annotations for classification, localization, and segmentation tasks. Our experimental results show that Foundation X achieves notable performance gains through extensive annotation utilization, excels in cross-dataset and cross-task learning, and further enhances performance in organ localization and segmentation tasks. All code and pretrained models are publicly accessible at~\href{https://github.com/jlianglab/Foundation_X}{GitHub.com/JLiangLab/Foundation\_X}.
\end{abstract}

\begin{table*}[!t]
  \footnotesize
  \centering
  \begin{tabular}{ p{4.5cm}@{} P{4cm}@{} P{4cm}@{} P{4cm}@{} }
    \toprule
    Dataset & Classification (Head Id) & Localization (Decoder Id) & Segmentation (Head Id) \\
    \midrule
    1.  CheXpert~\cite{irvin2019chexpert}          & \checkmark ($C_1$) & - & -  \\
    2.  NIH ChestX-ray14~\cite{wang2017chestx}  & \checkmark ($C_2$) & - & -  \\
    3.  VinDr-CXR~\cite{nguyen2022vindr}         & \checkmark ($C_3$) & - & -  \\
    4.  NIH Shenzhen CXR~\cite{jaeger2014two}  & \checkmark ($C_4$) & - & -  \\
    5.  MIMIC-II~\cite{johnson2019mimic}         & \checkmark ($C_5$) & - & -  \\
    6.  TBX11k~\cite{liu2020rethinking}            & \checkmark ($C_6$) & \checkmark ($L_1$) & -  \\
    7.  NODE21~\cite{sogancioglu2024nodule}            & \checkmark ($C_7$) & \checkmark ($L_2$) & -  \\
    8.  CANDID-PTX~\cite{feng2022automated}        & \checkmark ($C_8$) & \checkmark ($L_3$) & \checkmark ($S_1$)  \\
    9.  RSNA Pneumonia~\cite{radiological2018rsna}    & \checkmark ($C_9$) & \checkmark ($L_4$) & -  \\
    10. ChestX-Det~\cite{lian2021structure}        & \checkmark ($C_{10}$) & \checkmark ($L_5$) & \checkmark ($S_2$) \\
    11. SIIM-ACR ~\cite{dramsch2019siim}         & \checkmark ($C_{11}$) & \checkmark ($L_6$) & \checkmark ($S_3$) \\ 
    \midrule
    12. CheXmask VinDr-CXR~\cite{gaggion2024chexmask}  &  -  & \checkmark  & \checkmark \\
    13. VinDr-RibCXR~\cite{nguyen2021vindrRIB}    & - & - & \checkmark \\
    14. NIH Montgomery~\cite{jaeger2014two}           & - & - & \checkmark \\
    15. JSRT~\cite{van2006segmentation}         &  -  &     -     & \checkmark \\
    16. VinDr-CXR~\cite{nguyen2022vindr} & - & \checkmark & - \\
    \bottomrule
  \end{tabular}
  \caption{We pretrain our Foundation X model using 11 publicly available chest X-ray datasets, as shown in the first 11 datasets in the table. Although not every dataset contains all three types of annotations—classification, localization, and segmentation—we leverage all available annotations to maximize the model's learning potential. Among these datasets, all include classification ground truths, 6 provide localization bounding box annotations, and 3 offer segmentation masks for diseases. Furthermore, we utilize organ localization and segmentation datasets from VinDr-CXR, VinDr-RibCXR, NIH Montgomery, and JSRT for target task finetuning. Here, the organ segmentation masks for VinDr-CXR were sourced from the CheXmask database. We also finetuned VinDr-CXR with local labels for disease localization task.}
  \label{tab:pretrainingdatasets}
\end{table*}

\section{Introduction}
\label{sec:intro}
In computer vision, tasks like classification, localization, and segmentation are often handled independently. This approach can lead to inefficiencies and limit performance in complex, real-world applications. Isolated models miss the opportunity to leverage the rich, diverse information available when these tasks are integrated~\cite{zhao2023multi,ghiasi2021multi}. 
In medical imaging, datasets often contain different diseases annotated with image-level labels, disease-specific bounding boxes, and segmentation masks. For example, CheXpert~\cite{irvin2019chexpert} dataset has only image-level labels for classification, while TBX11K~\cite{liu2020rethinking} has both image-level labels and bounding boxes, and CANDID-PTX~\cite{feng2022automated} provides annotations for all three tasks. By integrating these types of annotations into a single model, we can achieve a deeper understanding of each dataset. We hypothesize that combining classification (to identify diseases), localization (to generate bounding boxes), and segmentation (to delineate boundaries) tasks within the same framework will improve image analysis. This, in turn, will lead to better diagnostic accuracy and more informed medical decisions. 
Therefore, developing an end-to-end framework that handles all tasks simultaneously would boost performance and enhance robustness by taking advantage of the semantic relationships between tasks~\cite{ghiasi2021multi,he2022x,islam2024seeking}.
However, this integration poses a significant challenge, as model tend to overfit to a single task during training, hindering generalization across multiple tasks. To overcome this issue, our research proposes a framework that tackles these tasks concurrently and serves as a foundation model. By training it on large-scale, diverse datasets and tasks, we aim to build a system capable of handling a broad range of real-world applications, improving both task-specific performance and generalizability. 
Such a model leverages the synergy between classification, localization, and segmentation, creating a versatile, robust system while maximizing annotation use, reducing costs, and enhancing efficiency in medical image analysis.
This leads us to our central research question:~\textit{How can we integrate classification, localization, and segmentation tasks within a single model to improve its performance and generalization ability, specifically in Chest X-ray image analysis?} In our research, we have chosen Chest X-rays (CXRs) because they are one of the most frequently used imaging modalities, and the availability of CXR data is substantial (see Table~\ref{tab:pretrainingdatasets}).

\begin{figure*}[!t]
    \centering
    \includegraphics[width=1\textwidth]{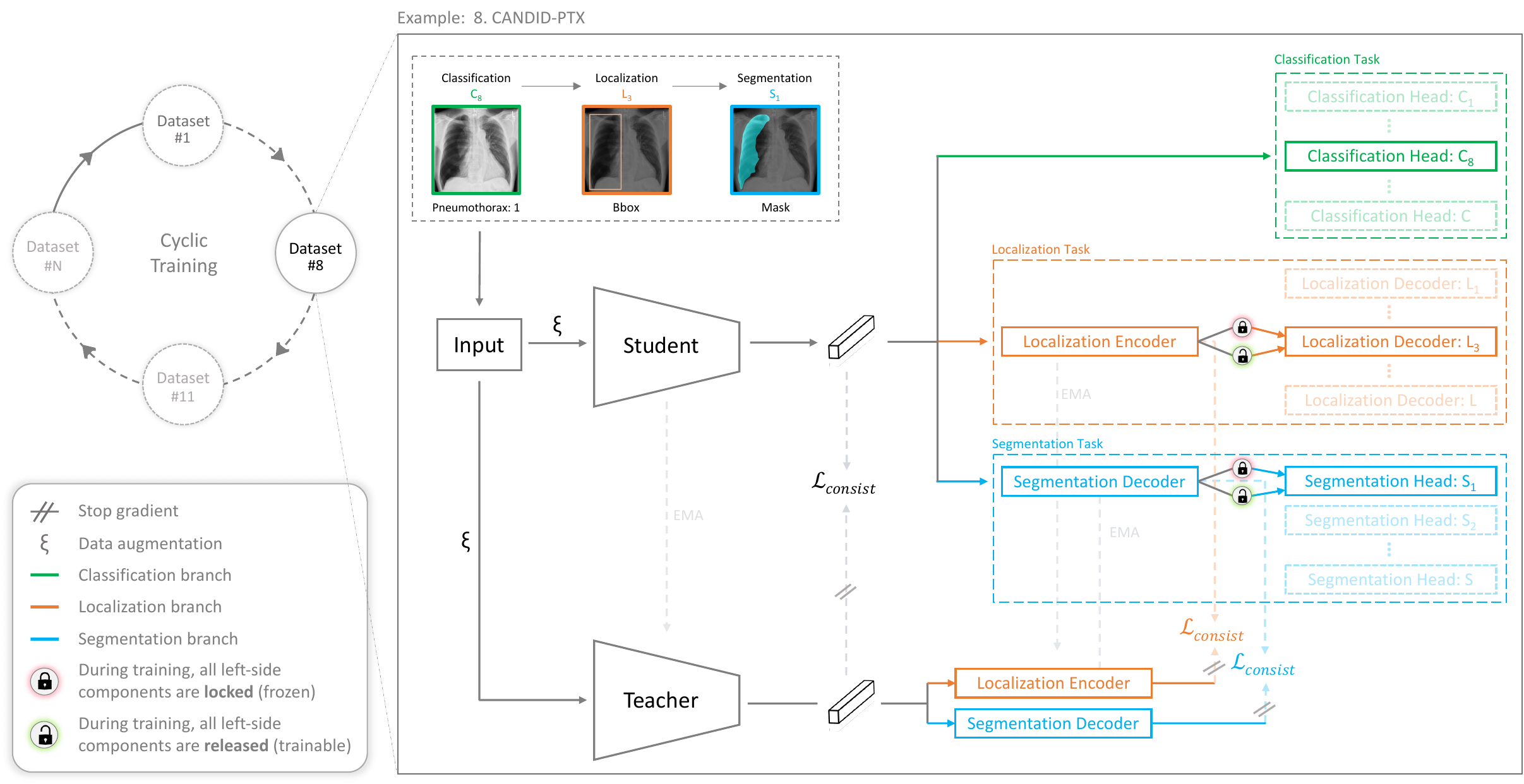}
    \caption{
    The proposed Foundation X model ({detailed in Sec.~\ref{sec:methods}}) can {utilize} multiple datasets (Dataset \#1 to \#11) for pretraining and can also incorporate additional datasets (Dataset \#N) dynamically into the pretraining process. The model is trained cyclically, processing each dataset sequentially. Each dataset may include one, two, or all three tasks: classification, localization, and segmentation. The figure illustrates the process with a dataset (e.g., 8. CANDID-PTX~\cite{feng2022automated}) containing all three types of ground truths. The process begins with the student model (Swin-B) extracting relevant features from the input dataset, which are then directed sequentially to the appropriate branch. First, for classification, features are processed through the classification head ($C_8$). Second, for localization, features pass through the localization encoder and corresponding localization decoder ($L_3$). Third, for segmentation, features are handled by the segmentation decoder and segmentation head ($S_1$). The model undergoes two-phase training for each task: lock mode with most layers frozen, followed by release mode with all layers trainable. 
    Additionally, the model uses a student-teacher learning paradigm. The teacher model, an identical copy of the student model, is updated after each epoch using an exponential moving average (EMA). We calculate the consistency loss ($L_{\text{const}}$) in three areas: extracted features from the backbone, features from the localization encoder, and features from the segmentation decoder. If a dataset contains only one or two types of ground truths, the model will skip the branch without the corresponding ground truth. The Foundation X model uses the Cyclic and Lock-Release pretraining strategies to enhance performance across tasks while preventing forgetting and avoiding task-specific overfitting.}
    \label{IntegratedModelFig}
\end{figure*}

To this extent, we have developed~\textbf{Foundation X}, an end-to-end model that integrates classification, localization, and segmentation tasks for medical imaging (illustrated in Figure~\ref{IntegratedModelFig}).
We hypothesize that sharing learned representations across tasks equips the model to better capture intricate patterns in medical images, leading to more reliable diagnostics. However, integrating classification, localization, and segmentation into a single model risks overfitting, especially during large-scale training. Foundation X tackles this by using a shared backbone and innovative pretraining strategies, ensuring balanced performance across diverse tasks.
To demonstrate the capability of Foundation X, we train the model on 11 datasets (Table~\ref{tab:pretrainingdatasets}) and further finetune it on additional target tasks, showcasing the model's potential.

In summary, our analysis shows that Foundation X achieves significant performance gains through extensive annotation usage (Table~\ref{tab:pretrainingXclslocseg}), excels in cross-dataset and cross-task learning (Figure~\ref{X_CrossDataset_CrossTask2}), and further enhances organ localization and segmentation (Table~\ref{tab:vindrcsrorgansegmentation_all},~\ref{tab:JSRT_full},~\ref{tab:JSRT_clavicle_fewshot}).
Through this work, we have made the following contributions:

\begin{enumerate}
\vspace{-0.65em}
\item {We develop and implement} Foundation X, an integrated model for classification, localization, and segmentation tasks in Chest X-ray images;
\vspace{-0.65em}
\item We propose a Lock-Release pretraining strategy to enhance the cyclic learning from multiple datasets, preventing task overfitting and ensuring balanced learning across tasks and datasets;
\vspace{-0.65em}
\item {We provide} comprehensive experimental results {to demonstrate} Foundation X's improved performance and generalization ability.
\end{enumerate}

\vspace{-0.6em}
\noindent Foundation X provides a versatile, end-to-end framework for handling classification, localization, and segmentation tasks in medical imaging. By leveraging shared knowledge across tasks, Foundation X enhances generalization, reduces overfitting, and maximizes annotation use, leading to more efficient data utilization. This approach allows the model to adapt to new tasks and datasets, making it valuable for continuous learning and real-world medical use.

\section{Related Work}
\label{sec:relatedwork}

Multitask learning mimics the human brain by performing tasks simultaneously with minimal supervision and simplifying cross-learning from different tasks. Significant attention has been given to multitask learning~\cite{caruana1997multitask} within the research community.  In recent years, researchers have focused on utilizing a single backbone with one or more decoders for multitask learning~\cite{malhotra2022multi,playout2019novel, Li_2023_CVPR, He_2017_ICCV, zhu2021deformabledetrdeformabletransformers}. However, the information shared between these unified or separate decoders was often ineffective, limiting performance and being restricted to either single-task multi-source or single-source multitask objectives. Researchers have explored various approaches to address and improve these learning methods. One approach is OmniFM-DR~\cite{xu2023learning}, which uses the text encoder in parallel with a backbone and along with multiple unified decoders to perform multitask learning. However, the expert annotations are modified to create the input text features for the encoder and the unified multitask decoder architecture. 
Other learning methods~\cite{li2021referring,lu2022unified,wang2022ofa} involve using an image backbone and context encoder, then feeding these features into a visual-lingual encoder and decoder.
However, these one-stage model uses multiple heads for multitask learning, limited to either a text encoder or multiple backbones for the encoder-decoder architecture.

On the other hand, the framework X-Learner~\cite{he2022x} consists of two stages: Expansion, where multiple sub-backbones are learned and interconnected to reduce task interference, and Squeeze, where the expanded backbone is condensed into a normal-sized one for effective downstream transfer.
However, the multiple sub-backbones may not efficiently capture information between tasks, potentially leading to suboptimal performance on some tasks. Additionally, the lack of an open code-base limits opportunities for comparative studies and comprehensive evaluations.

For techniques to aggregate dataset annotations, Universal {Object} {Detection} with {Large} {Vision} {Model}~\cite{lin_universal_2024} proposes to modify the existing hierarchy of the dataset labels to create a unified label space and incorporate a hierarchical loss suppression technique to efficiently calculate losses for the taxonomies in the labels. Our Foundation X framework differs from~\cite{lin_universal_2024} by aggregating multiple datasets without any modifications to the label taxonomies or the classes themselves through the use of multiple task-heads. Each task-head in Foundation X is used for a single dataset and vision task. 
The work, {ViM} ({Vision} {Middleware} for {Unified} {Downstream} {Transferring})~\cite{feng_vim_2023}, takes on accommodating multiple different tasks in a single model by introducing a new paradigm. 
The design proposed by~\cite{feng_vim_2023} involves using a fully frozen backbone pretrained on a large dataset for an upstream task. It then combines various ViM modules (adapters) trained on datasets such as COCO~\cite{lin2015microsoftcococommonobjects}, Objects365~\cite{shao_objects365_2019}, and others, to finally finetune on the target downstream dataset.
Our method contrasts~\cite{feng_vim_2023} by having a simpler but effective architecture that inherently learns generalizable representations from multiple datasets and tasks. 

Foundation Ark~\cite{ma2023foundation} first introduced the idea of accruing and reusing knowledge embedded in the expert annotations from numerous datasets cyclically, but it focused solely on the classification task. In contrast, Foundation X focuses on encompassing diverse tasks, including classification, localization, and segmentation, via a Lock-Release strategy.

\vspace{-0.3em}
\section{Method}
\label{sec:methods}
\subsection{Foundation X: Integrating Classification, Localization and Segmentation}
Our Foundation X aims to integrate classification, localization, and segmentation tasks to train a powerful, robust, and versatile model. This integrated approach allows the model to develop a comprehensive understanding of images, enhancing its performance across all tasks. By leveraging the synergistic effect of these tasks, we can improve overall diagnostic accuracy and maximize the utilization of available annotated data, thereby saving the annotation costs. To achieve this, we designed Foundation X with a shared backbone to learn general and complementary knowledge across tasks, and separate branches to address the specific needs of classification, localization, and segmentation. 
This separation enables independent optimization and finetuning of each branch, giving Foundation X the flexibility and scalability to add new tasks and datasets while maintaining high performance across diverse applications.

As shown in Figure~\ref{IntegratedModelFig}, the backbone of Foundation X serves as a knowledge encoder, extracting common features for various tasks performed by three branches: the classification branch with multiple heads, the localization branch with an encoder and multiple decoders, and the segmentation branch with a decoder and multiple heads. Each head or localization decoder corresponds to a specific dataset, as listed in Table~\ref{tab:pretrainingdatasets}.
For the classification branch, we implement the classification head with a linear classifier. 
For the localization branch, we utilize the localization encoder and decoder from DINO (DETR with Improved deNoising anchOr boxes)~\cite{zhang2022dino}, with modifications to accommodate multiple datasets by using multiple localization decoders. This design ensures that each localization dataset has a dedicated decoder, allowing the model to effectively handle multiple tasks.
For the segmentation branch, we utilize the UperNet decoder~\cite{xiao2018unifiedUperNet} and add multiple segmentation heads atop it to handle different segmentation datasets. 
The design of multiple heads and localization decoders enhances the flexibility of Foundation X, allowing it to seamlessly add new tasks and making it scalable for task expansion.

\begin{table}[t]
    \centering
    \footnotesize
    \begin{tabular}{ p{2.5cm}@{} p{5cm}@{} }
    \textbf{Configuration} & \textbf{Utilization of Branch} \\
        \midrule
        Foundation X-C   & Only \textbf{C}lassification \\
        Foundation X-L   & Only \textbf{L}ocalization \\
        Foundation X-S   & Only \textbf{S}egmentation \\
        Foundation X-CL  & \textbf{C}lassification and \textbf{L}ocalization \\
        Foundation X-LS  & \textbf{L}ocalization and \textbf{S}egmentation \\
        Foundation X-CLS & \textbf{C}lassification, \textbf{L}ocalization \& \textbf{S}egmentation \\
    \end{tabular}
    \caption{Differences between Foundation X model configurations.}
    \label{tab:foundationx_configurations}
\end{table}

\subsection{Cyclic Pretraining Strategy}

Training a single model using diverse datasets with inconsistent or heterogeneous annotations is challenging. The model must learn from various types of information and integrate them into a cohesive understanding. For example, the model needs to extract high-level representations of the entire image for classification, identify specific regions of interest for localization, and delineate the precise boundaries of these regions for segmentation. Even for the same type of task, datasets created at different institutions tend to be annotated differently, further complicating the learning process~\cite{ma2023foundation}. 

To address these challenges, we employ the Cyclic pretraining strategy from Foundation Ark~\cite{ma2023foundation}. This strategy enables the model to learn from multiple tasks by revisiting each one in every training round, thereby reinforcing the learning process and preventing the model from forgetting previously acquired knowledge. Benefiting from the Cyclic pretraining, Foundation X avoids the issue of catastrophic forgetting~\cite{kemker2018measuring, ma2023foundation}, where performance on earlier tasks significantly degrades when new tasks are introduced. This approach enables the model to achieve more robust and generalized performance, enhancing its effectiveness across multiple datasets and tasks.

\subsection{Lock-Release Pretraining Strategy}
Another challenge when training a model on diverse datasets and tasks is ensuring it maintains good generalizability across all, without overfitting towards any single dataset or task. The model must be capable of learning from various tasks, understanding heterogeneous annotations, and integrating sophisticated domain knowledge into a cohesive framework that performs well across different tasks. However, due to the varying number of training samples and the differing difficulty levels of each task, the model's learning speed can vary.

To balance the learning process for each task, we have developed a Lock-Release pretraining strategy for localization and segmentation tasks. Initially, the model is trained in the \textit{lock} mode, where most early layers are frozen and only a few upper layers are trainable. Specifically, for localization and segmentation tasks, only the localization decoder and segmentation head are trainable, respectively. This mode focuses on finetuning higher-level features specific to the task while preserving general features learned from other datasets. Subsequently, training switches to the \textit{release} mode, where all layers are made trainable, allowing for full adaptation and refinement. During the \textit{lock} mode training, only half of the dataset is used to prevent early overfitting, while the full dataset is utilized during the \textit{release} mode training to ensure comprehensive learning. This Lock-Release strategy helps prevent the model from overfitting too heavily towards one task when exposed to multiple tasks and datasets, ensuring a more balanced learning process.

We tried the Lock-Release strategy for classification but found it ineffective, so it was not applied. Unlike localization and segmentation, which have more parameters to tune, classification relies on lightweight heads, making Lock-Release less impactful.

\subsection{Student-Teacher Learning Paradigm}
To further mitigate the issue of forgetting and prevent overfitting towards any single task, thereby balancing and stabilizing the learning process across diverse tasks and datasets, we introduce the student-teacher learning paradigm in Foundation X. The teacher model uses the same architecture as the student model, and both are initialized with the same weights. The student model is updated through standard training processes, while the teacher model is updated using an exponential moving average (EMA)~\cite{tarvainen2017mean} based on the student's learning at the end of each epoch. Furthermore, as shown in Figure~\ref{IntegratedModelFig}, We incorporate a consistency loss for the features from the backbone, the localization encoder, and the segmentation decoder between the student and teacher models. This consistency loss ensures that the features learned by the student model remain aligned with those of the teacher model, promoting stability and improved performance during training. The student-teacher learning paradigm enhances generalization across classification, localization, and segmentation tasks, resulting in improved performance and robustness on various tasks. After pretraining, the teacher model from Foundation X will be used for the downstream tasks.

\section{Experiments and Results}
\label{sec:experiments&results}
\noindent\textbf{Pretraining Foundation X.} Foundation X is pretrained on 11 datasets (see Table~\ref{tab:pretrainingdatasets}) encompassing three common medical tasks: \textit{disease} classification, localization, and segmentation. Among these datasets, all provide disease classification, two include localization and three offer segmentation mask annotations. Since CANDID-PTX~\cite{feng2022automated} and SIIM-ACR Pneumothorax~\cite{dramsch2019siim} are annotated only with disease masks, we derive localization bounding boxes from them, resulting in a total of four localization tasks. We use the official dataset splits when available and perform random splits (70\% train, 10\% val, 20\% test) for those without.

\noindent\textbf{Finetuning Foundation X.} We collect an additional 5 publicly available chest X-ray datasets to finetune Foundation X on \textit{organ} segmentation and localization tasks. CheXmask~\cite{gaggion2024chexmask} offers a comprehensive collection of uniformly annotated chest radiographs compiled from five public sources: ChestX-ray8, Chexpert, MIMIC-CXR-JPG, Padchest, and VinDr-CXR. It includes segmentation masks for the heart, left lung, and right lung. From these provided segmentation masks, we derive localization bounding boxes for the same organs. We specifically utilize the VinDr-CXR portion of the dataset to localize and segment the organs. 
Furthermore, we use the NIH Montgomery~\cite{jaeger2014two}, VinDr-RibCXR~\cite{nguyen2021vindrRIB}, and JSRT~\cite{shiraishi2000development} datasets for organ segmentation in our Foundation X model. NIH Montgomery provides lung masks, VinDr-RibCXR provides rib masks, and JSRT provides heart, lung, and clavicle masks.

\begin{table}[h]
  \centering
  \footnotesize
  \begin{tabular}{ p{1.40cm}@{} P{1.35cm}@{} P{1.35cm}@{} P{2cm}@{} P{2cm}@{} }
    \toprule
    Dataset VinDr-CXR & Baseline Loc. & Baseline Seg. & \multicolumn{2}{c}{Foundation X-LS} \\
     & \scriptsize[mAP40\%] & \scriptsize[Dice\%] & \scriptsize [mAP40\%] & \scriptsize[Dice\%] \\  
    \midrule
    Heart      & 80.17 & 95.82 & \textbf{ 88.41 }{\footnotesize\NIgreen{$\uparrow8.24$}} & \textbf{ 96.15 }{\footnotesize\NIgreen{$\uparrow0.33$}} \\
    Left Lung  & 90.72 & 97.46 & \textbf{ 95.58 }{\footnotesize\NIgreen{$\uparrow4.86$}} & \textbf{ 97.57 }{\footnotesize\NIgreen{$\uparrow0.11$}} \\
    Right Lung & 92.42 & 98.03 & \textbf{ 96.78 }{\footnotesize\NIgreen{$\uparrow4.36$}} & \textbf{ 98.13 }{\footnotesize\NIgreen{$\uparrow0.10$}} \\
    \bottomrule
  \end{tabular}
  \caption{
  Baseline localization and segmentation models are trained separately using DINO~\cite{zhang2022dino} and UperNet~\cite{xiao2018unifiedUperNet} to localize and segment the heart, left lung, and right lung in the VinDr-CXR dataset, each employing a single head for three classes. In contrast, our Foundation X-LS model handles both tasks together in pretraining, showing enhanced performance. The green arrow highlights Foundation X's performance improvements over the baselines.
  }
  \vspace{-0.5em}
  \label{tab:vindrcsrorgansegmentation_all}
\end{table}

\begin{table*}
  \footnotesize
  \centering
  \begin{threeparttable}
  \begin{tabular}{ p{2.5cm}@{} P{1.85cm}@{} P{1.85cm}@{} P{2cm}   P{2.5cm} P{2.5cm} P{2.5cm} }
    \toprule
    Dataset & Baseline Cls.$^{\dagger}$ & Baseline Loc.$^{\dagger}$ & Baseline Seg.$^{\dagger}$ & \multicolumn{3}{c}{Foundation X-CLS} \\
            & \scriptsize[AUC\%] & \scriptsize[mAP40\%] & \scriptsize[Dice\%] & \scriptsize[AUC\%] & \scriptsize[mAP40\%] & \scriptsize[Dice\%] \\   
    \midrule
    CheXpert         & 90.03\footnotesize$\pm0.48$ & -     & -     & \textbf{90.64 }{\footnotesize\NIgreen{$\uparrow0.61$}} & - & - \\
    NIH ChestX-ray14 & 83.05\footnotesize$\pm0.09$ & -     & -     & {82.95 }({\textbf{83.35}\footnotesize*}{\footnotesize\NIgreen{$\uparrow0.30$}}) & - & - \\
    VinDr-CXR        & 95.07\footnotesize$\pm0.15$ & -     & -     & \textbf{95.85 }{\footnotesize\NIgreen{$\uparrow0.78$}} & - & - \\
    NIH Shenzhen CXR & 98.99\footnotesize$\pm0.16$ & -     & -     & \textbf{99.64 }{\footnotesize\NIgreen{$\uparrow0.65$}} & - & - \\
    MIMIC-II         & 79.12\footnotesize$\pm0.16$ & -     & -     & {78.94 }{\footnotesize\NIred{$\downarrow0.18$}} & - & - \\
    
    TBX11K           & 99.89\footnotesize$\pm0.06$ & 78.08\footnotesize$\pm0.81$ & -     & \textbf{99.95 }{\footnotesize\NIgreen{$\uparrow0.06$}} & {78.38 (\textbf{81.80}\footnotesize*}{\footnotesize\NIgreen{$\uparrow3.72$}}) & - \\
    NODE21           & 99.35\footnotesize$\pm0.45$ & 37.78\footnotesize$\pm2.67$ & -     & \textbf{99.68 }{\footnotesize\NIgreen{$\uparrow0.33$}} & \textbf{46.57 }{\footnotesize\NIgreen{$\uparrow8.79$}} & - \\
    CANDID-PTX       & 72.61\footnotesize$\pm0.57$ & 50.51\footnotesize$\pm1.36$ & 86.36 & \textbf{73.86 }{\footnotesize\NIgreen{$\uparrow1.25$}} & \textbf{54.14 }{\footnotesize\NIgreen{$\uparrow3.63$}} & \textbf{89.81 }{\footnotesize\NIgreen{$\uparrow3.45$}} \\
    RSNA Pneumonia   & 88.87\footnotesize$\pm0.21$ & 20.83\footnotesize$\pm0.54$ & -     & \textbf{89.88 }{\footnotesize\NIgreen{$\uparrow1.01$}} & \textbf{27.44 }{\footnotesize\NIgreen{$\uparrow6.61$}} & - \\
    ChestX-Det       & 88.17\footnotesize$\pm0.33$ & 38.12\footnotesize$\pm0.50$ & \textbf{79.33} & {85.07 }({\textbf{89.89}\footnotesize*}{\footnotesize\NIgreen{$\uparrow1.72$}}) & {37.77 }({\textbf{43.98}\footnotesize*}{\footnotesize\NIgreen{$\uparrow5.86$}})  & {64.49 }({79.17\footnotesize*}{\footnotesize\NIred{$\downarrow0.16$}}) \\
    SIIM-ACR         & 95.01\footnotesize$\pm0.16$ & 28.56\footnotesize$\pm0.94$ & 81.92 & \textbf{96.44 }{\footnotesize\NIgreen{$\uparrow1.43$}} & \textbf{34.59 }{\footnotesize\NIgreen{$\uparrow6.03$}} & \textbf{83.65 }{\footnotesize\NIgreen{$\uparrow1.73$}} \\
    \bottomrule
  \end{tabular}
    \begin{tablenotes}
    \item \footnotesize$^{\dagger}$ All baselines use Swin-B~\cite{liu2021swin} as the backbone with Ark-6~\cite{ma2023foundation} pretrained weights. The classification baseline uses only Swin-B, the localization baseline uses Swin-B + DINO~\cite{zhang2022dino}, and the segmentation baseline uses Swin-B + UperNet~\cite{xiao2018unifiedUperNet}.
    \item \footnotesize$^{*}$ Values inside parentheses indicate the finetuning results of Foundation X, while the preceding values represent the pretraining results.
    \end{tablenotes}
    \end{threeparttable}
  \caption{Performance Comparison of Foundation X and Baseline Models. The table presents the results from pretraining Foundation X, compared to baseline models, across 11 datasets encompassing 20 tasks (see Table~\ref{tab:pretrainingdatasets}). Foundation X is trained sequentially on these tasks using  the Cyclic and Lock-Release pretraining strategies, which helps it generalize efficiently and retain knowledge of previous tasks. The results indicate that Foundation X outperforms most of the baselines, which are trained individually on specific datasets and tasks. This highlights the effectiveness of the integrated multitask learning approach in improving model performance and generalization ability. The arrow shows Foundation X’s performance gain/loss compared with the baseline performance. 
  }
  \vspace{-0.1em}
  \label{tab:pretrainingXclslocseg}
\end{table*}

\subsection{Foundation X achieves performance gains through extensive annotation utilization}
\noindent\underline{\textit{Experimental Setup}}: We pretrain Foundation X on a large-scale dataset collection, utilizing the first 11 datasets from Table~\ref{tab:pretrainingdatasets}, which encompass 20 tasks ($C_1$-$C_{11}$ for classification, $L_1$-$L_6$ for localization, and $S_1$-$S_3$ for segmentation). We initialize the Swin-B backbone with Ark-6~\cite{ma2023foundation} pretrained weights. For each dataset, we pretrain Foundation X on all tasks sequentially. For example, epoch $x$ is dedicated to classification, $x+1$ to localization, and $x+2$ to segmentation. If a dataset lacks annotations for a specific task, that task is simply skipped. We define one \textit{cycle} as the model completing training on all 20 tasks listed in Table~\ref{tab:pretrainingdatasets}. During pretraining, we employ Student-Teacher learning paradigm, Cyclic and Lock-Release pretraining strategies. This approach ensures comprehensive exposure to all tasks and datasets, promoting better generalization and robust performance in medical imaging.

\vspace{+0.3em}
\noindent\underline{\textit{Results and Analysis}}: 
Using Cyclic and Lock-Release pretraining strategies, Foundation X achieves significantly better performance on most tasks with large-scale pretraining. As shown in Table~\ref{tab:pretrainingXclslocseg}, it outperforms baseline models that are independently trained on specific datasets and tasks.
Notably, due to our superior pretraining strategy, Foundation X does not exhibit overfitting toward a single task, resulting in consistent improvement across various datasets and tasks.
In contrast, baseline models are fully finetuned on individual datasets, giving them an advantage in single-task performance. Consequently, during Foundation X's large-scale pretraining, it is expected that some tasks may face difficulties due to the need to balance learning across diverse datasets and tasks. Despite this challenge, Foundation X achieves comparable or better results than the baselines in most cases. To address underperforming tasks during pretraining, we further finetune Foundation X from the latest checkpoint, achieving better results than the baselines. As shown in Table~\ref{tab:pretrainingXclslocseg}, the NIH ChestX-ray14 disease classification task improves to 83.35\%, the localization performance on TBX11K increases to 81.80\%, the ChestX-Det classification raises to 89.89\%, and ChestX-Det localization raises to 43.98\%. However, finetuning for the ChestX-Det segmentation task (79.17\%) showed limited improvement, likely due to the low class-to-data ratio.

\begin{table}[!t]
  \centering
  \footnotesize
  \begin{threeparttable}
  \begin{tabular}{ p{2.15cm}@{} P{1.55cm}@{} P{1.65cm}@{} P{2.3cm}@{} }
    \toprule
    Dataset    & Ark$^{\dagger}$\footnotesize\cite{ma2023foundation} & POPAR$^{\ddagger}$\footnotesize\cite{pang2022popar} & Foundation X-LS \\
               & \scriptsize[Dice\%] & \scriptsize[Dice\%] & \scriptsize[Dice\%] \\
    \toprule
    JSRT-Heart    & 94.62\footnotesize$\pm0.16$ & \underline{94.64}\footnotesize$\pm0.21$ & \textbf{ 95.42 }{\footnotesize$\pm0.02$\footnotesize\NIgreen{$\uparrow0.78$}} \\
    JSRT-Lung     & 97.48\footnotesize$\pm0.06$ & \underline{97.71}\footnotesize$\pm0.07$ & \textbf{ 98.04 }{\footnotesize$\pm0.04$\footnotesize\NIgreen{$\uparrow0.33$}} \\
    JSRT-Clavicle & 90.05\footnotesize$\pm0.15$ & \underline{90.18}\footnotesize$\pm0.18$ & \textbf{ 91.17 }{\footnotesize$\pm0.34$\footnotesize\NIgreen{$\uparrow0.99$}} \\
     
    NIH Montgomery     & 97.68\footnotesize$\pm0.03$ & \underline{97.78}\footnotesize$\pm0.05$ & \textbf{ 98.29 }{\footnotesize$\pm0.02$\footnotesize\NIgreen{$\uparrow0.51$}} \\
    
    VinDr-RibCXR    & \underline{63.96}\footnotesize$\pm0.30$ & 61.17\footnotesize$\pm0.40$ & \textbf{ 71.12 }{\footnotesize$\pm0.56$\footnotesize\NIgreen{$\uparrow7.16$}} \\
    \toprule
    
  \end{tabular}
    \begin{tablenotes}
    \item \footnotesize$^{\dagger}$ We adopt this performance reported by the original authors~\cite{ma2023foundation}. 
    \item \footnotesize$^{\ddagger}$ POPAR~\cite{pang2022popar} is finetuned for the baselines.
    \end{tablenotes}
    \end{threeparttable}
  \caption{We initialize Foundation X-S with weights from the Foundation X-LS model, trained on VinDr-CXR for organ localization and segmentation. After finetuning on other target tasks, Foundation X-S shows notable performance gains over Ark and POPAR. The green arrow indicates its improvements over the second-best method (underlined).}
  \vspace{-1.9em}
  \label{tab:JSRT_full}
\end{table}

\subsection{Foundation X Enhances Performance for Organ Loc. and Seg.}
\noindent\underline{\textit{Experimental Setup}}: We pretrain Foundation X exclusively on organ localization and segmentation tasks using the VinDr-CXR dataset. We first evenly divide the official training split, resulting in 7,500 non-overlapping images for both localization and segmentation pretraining tasks. Additionally, we further divide the dataset into three non-overlapping subsets for three specific organs (heart, left lung, and right lung). This careful partitioning ensures a rigorous assessment of Foundation X's performance in localizing and segmenting these three organs independently. To establish baseline performance, we train DINO~\cite{zhang2022dino} for localization and UperNet~\cite{xiao2018unifiedUperNet} for segmentation on the aforementioned tasks separately. To demonstrate Foundation X's superior pretraining strategy on localization and segmentation tasks, we pretrain Foundation X using both localization and segmentation branches along with the Cyclic and Lock-Release pretraining strategies. We refer to the resulting model as \textit{Foundation X-LS} (Table~\ref{tab:vindrcsrorgansegmentation_all}) since it is trained exclusively on \textbf{L}ocalization and \textbf{S}egmentation tasks. To demonstrate that Foundation X-LS provides superior fine-grained features, we further finetune the model on three organ segmentation tasks: JSRT, NIH Montgomery, and VinDr-RibCXR. Lastly, we evaluate the model's effectiveness in few-shot learning setups using the JSRT-clavicle dataset. 
Here, we compare our results against two pretraining baselines: Ark~\cite{ma2023foundation} and POPAR~\cite{pang2022popar}. Ark utilizes the Swin-B backbone pretrained on 6 chest X-ray datasets in a supervised setup using the cyclic pretraining strategy. POPAR, a self-supervised method, also employs Swin-B pretrained on NIH ChestX-ray14 images, leveraging consistent and recurrent anatomical patterns in medical images to learn patch-level spatial relationships and fine-grained appearance features. For comparison on the segmentation task, we build the baseline using Swin-B + UperNet, where the Swin-B backbone is initialized with Ark-6 or POPAR pretrained weights.

\vspace{+0.2em}
\noindent\underline{\textit{Results and Analysis}}: 
{ As shown in Table~\ref{tab:vindrcsrorgansegmentation_all}, compared to the two baseline methods, DINO and UperNet, which are independently trained for specific tasks, the Foundation X-LS model—pretrained by sequentially incorporating both localization and segmentation—achieves the best performance. Specifically, during pretraining,}
we observe substantial gains in localization, with performance increases of 8.24\% for the heart, 4.86\% for the left lung, and 4.36\% for the right lung. Additionally, segmentation tasks show improvements of 0.33\%, 0.11\%, and 0.10\% for the heart, left lung, and right lung, respectively.
Table~\ref{tab:JSRT_full} demonstrates the Foundation X-LS model's finetuning performance on the individual task.
When finetuning on the JSRT dataset, we observe performance gains of 0.78\% for heart, 0.33\% for lung, and 0.99\% for clavicle segmentation. Additionally, we note gains of 0.51\% for the NIH Montgomery and 7.16\% for the VinDr-RibCXR segmentation task. In a few-shot setup for clavicle segmentation on the JSRT dataset (Table~\ref{tab:JSRT_clavicle_fewshot}), Foundation X consistently outperforms Ark and POPAR across all training sample sizes.

\vspace{-0.6em}
\begin{table}[h]
  \centering
  \footnotesize
  \begin{tabular}{ P{1.5cm}@{} P{1.55cm}@{} P{1.55cm}@{} P{2.6cm}@{} }
    \toprule
    Training & Ark\footnotesize~\cite{ma2023foundation} & POPAR\footnotesize~\cite{pang2022popar} & Foundation X-LS \\
    Samples  & \scriptsize[Dice\%] & \scriptsize[Dice\%] & \scriptsize[Dice\%] \\   
    \midrule
    24 & \underline{86.32} & 86.14 & \textbf{ 88.81 }{\footnotesize\NIgreen{$\uparrow2.49$}} \\
    20 & 84.87 & \underline{86.27} & \textbf{ 88.23 }{\footnotesize\NIgreen{$\uparrow1.96$}} \\
    15 & \underline{84.73} & 83.23 & \textbf{ 86.65 }{\footnotesize\NIgreen{$\uparrow1.92$}} \\
    12 & 80.82 & \underline{81.46} & \textbf{ 85.89 }{\footnotesize\NIgreen{$\uparrow4.43$}} \\
    6  & \underline{82.71} & 79.03 & \textbf{ 83.03 }{\footnotesize\NIgreen{$\uparrow0.32$}} \\
    3  & \underline{74.98} & 70.68 & \textbf{ 78.18 }{\footnotesize\NIgreen{$\uparrow3.20$}} \\
    \bottomrule
  \end{tabular}
  \caption{ 
  We finetune Foundation X-LS which is trained on VinDr-CXR for organ localization and segmentation. We then finetune it for JSRT clavicle segmentation in a few-shot learning setup. Our results consistently show that Foundation X-S outperforms Ark and POPAR, with the green arrow indicating performance boosts over the second-best method (underlined).
  }
  \label{tab:JSRT_clavicle_fewshot}
  \vspace{-0.5em}
\end{table}

\begin{figure*}[t]
    \centering
    \includegraphics[width=0.95\textwidth]{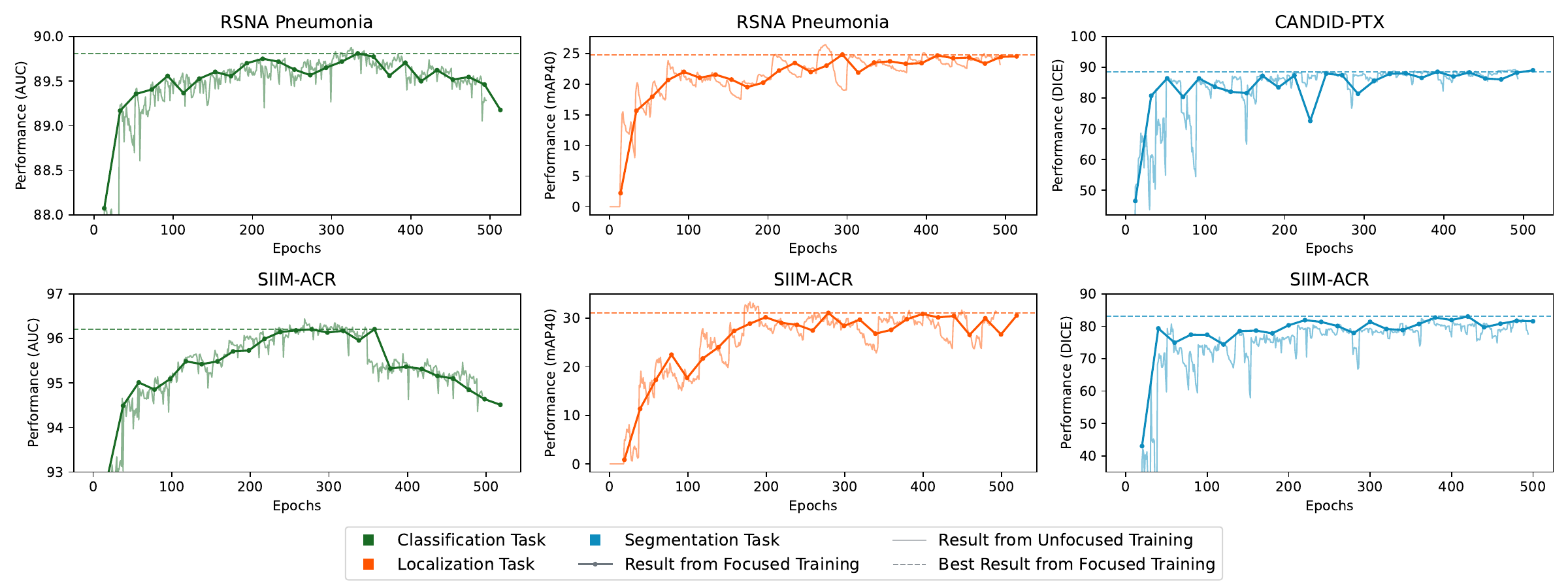}
\caption{Cross-Dataset \& Cross-Task Learning Analysis. The figure demonstrates the performance trends of Foundation X across multiple datasets for both focused and unfocused training scenarios. 
Focused training refers to scenarios where the model is explicitly trained on the specific dataset being evaluated. In contrast, unfocused training refers to scenarios where the model is trained on other datasets and not directly on the one being evaluated.
The green, orange, and blue lines represent classification, localization, and segmentation tasks, respectively. Dark-colored lines indicate focused training results, while light-colored lines show unfocused training results. Dashed lines represent the best testing outcomes from focused training. 
In some cases, unfocused training surpasses focused training, highlighting the benefits of cross-task and cross-dataset learning in enhancing Foundation X's capabilities. The model efficiently generalizes, retains knowledge of previous tasks, and avoids overfitting during pretraining.}
    \label{X_CrossDataset_CrossTask2}
\end{figure*}

\subsection{Foundation X maximizes performance with cross-dataset and cross-task learning}
\noindent\underline{\textit{Experimental Setup}}: 
We assess the generalizability and effectiveness of the Foundation X model across datasets and tasks. Specifically, We evaluate 1) how well the Foundation X model pretrained on one dataset performs on another, and 2) how the Foundation X model performs on two other tasks when pretrained by the third task within the same dataset. During the pretraining of Foundation X, we evaluate the model on the testing sets of all tasks across all datasets after each epoch, generating a series of Epoch V.S. Performance graphs. Our goal is to observe how testing performance changes over time when the model is trained on the same dataset (focused training) compared to when it is trained on other datasets (unfocused training). Figure~\ref{X_CrossDataset_CrossTask2} presents the results from focused training, unfocused training, and the best results from focused training for three selected datasets (RSNA Pneumonia, SIIM-ACR, and CANDID-PTX). Comprehensive plots for all datasets with multiple tasks are included in the supplementary document.

\vspace{+0.4em}
\noindent\underline{\textit{Results and Analysis}}: Figure~\ref{X_CrossDataset_CrossTask2} demonstrates positive performance trends across the datasets for both focused and unfocused training on each dataset. This indicates that the Foundation X model can effectively generalize and improve performance even without explicit training on specific tasks of the evaluated dataset.
During unfocused training (light-colored line), the performance dips are common initially, but improvement is typically observed over time. In all cases shown in Figure~\ref{X_CrossDataset_CrossTask2}, the results from unfocused training do not drift away from the task, indicating that the model can generalize efficiently and retain knowledge of previous tasks due to the Student-Teacher learning paradigm, Cyclic and Lock-Release pretraining strategies. The model performs consistently across all datasets and tasks. In some cases, unfocused training even outperforms focused training, highlighting the benefits of cross-task and cross-dataset learning in Foundation X.

\section{Discussion} \label{sec:discussion}
\begin{figure}[h!]
    \centering
    \includegraphics[width=0.97\columnwidth]{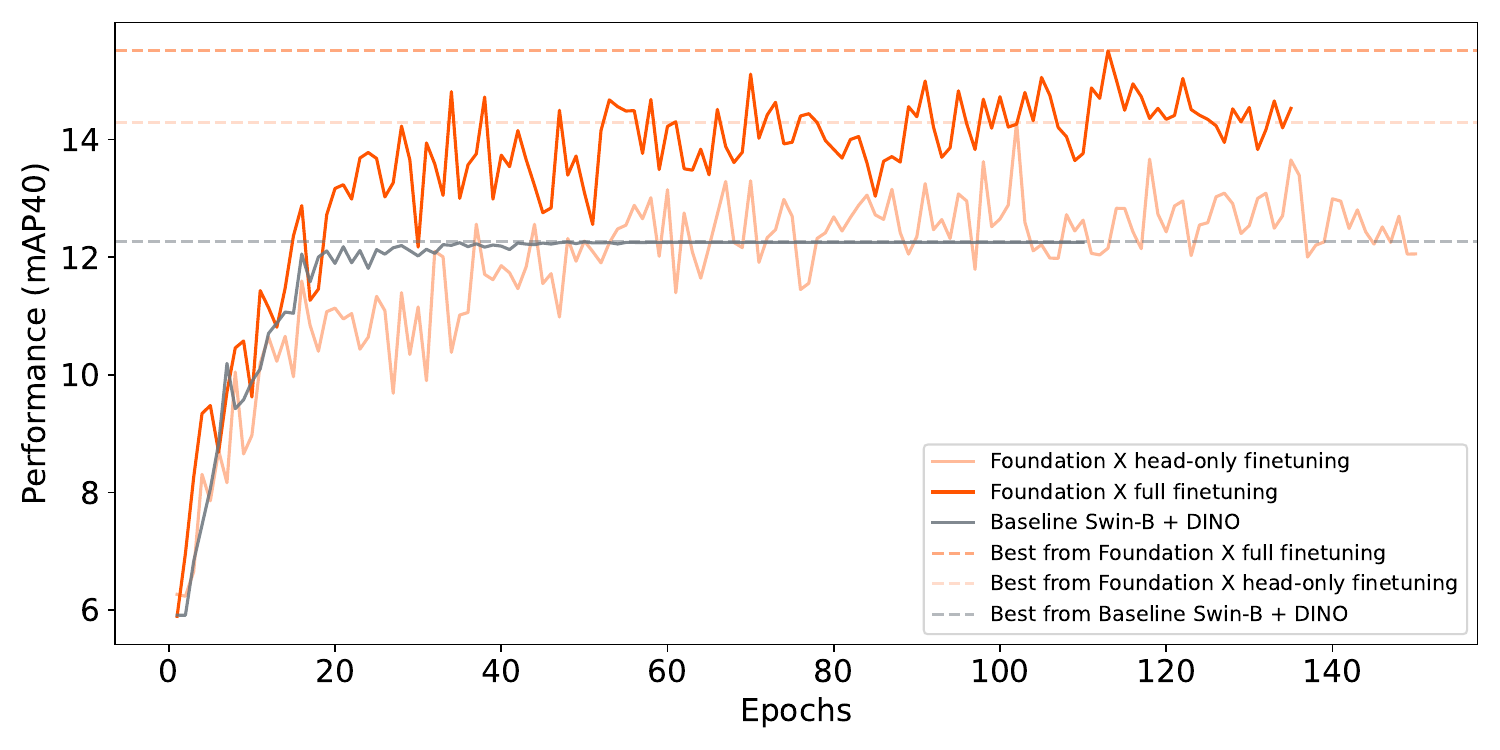}
    \caption{Full finetuning of Foundation X outperforms both head-only finetuning Foundation X and the baseline Swin-B + DINO model with Ark-6~\cite{ma2023foundation} initialized backbone weights. All three settings followed the same hyperparameters as mentioned in the supplementary material.} 
    \label{fig:head_only_ft_vs_baseline_fig}
    \vspace{-1em}
\end{figure}

Foundation X aims to develop a model that outperforms task-specific models by collaboratively learning from multiple datasets and tasks. Moreover, finetuning is required if a task-specific model needs to adapt to other tasks, such as adapting a classification pretrained model to a localization task. Our Foundation X model addresses evolving diagnostic task requirements through easy and quick finetuning. As observed in Figure~\ref{fig:head_only_ft_vs_baseline_fig}, finetuning Foundation X that is pretrained only on the classification task of the VinDr-CXR~\cite{nguyen2022vindr} dataset, amongst other pretraining datasets and tasks (Refer to Table \ref{tab:pretrainingdatasets}), achieves superior localization performance compared to the baseline model for the same dataset. We attribute this performance gain to the knowledge gathered from the Cyclic and Lock-Release pretraining strategies across all tasks and datasets. 

To further assess our Foundation X, we finetune the latest checkpoint from Foundation X-CLS using full finetuning, head-only finetuning and compare it with the localization baseline for the VinDr-CXR~\cite{nguyen2022vindr} localization dataset. 
In full finetuning, the model is trained on the VinDr-CXR localization dataset with a randomly initialized localization decoder. In head-only finetuning, the backbone and localization encoder are frozen, leaving only the new task head (localization decoder) trainable, with $\approx9.6$ million parameters.
While it is challenging to train a model with most of its layers frozen, head-only finetuning achieved $12.27\%$ mAP40 at the 49th epoch, where the baseline achieved its best result of $12.26\%$ mAP40 (see Figure~\ref{fig:head_only_ft_vs_baseline_fig}). The head-only finetuning of Foundation X, moreover, has significantly fewer trainable parameters than the baseline model (107M) but outperforms the latter by achieving its best result of $14.29\%$ mAP40 at the 101st epoch. Full finetuning of Foundation X, on the other hand, surpassed both head-only finetuning and baseline at epoch 49 with $13.10\%$ mAP40 and went on to achieve the best result of $15.51\%$ mAP40 at the 112th epoch. 
This reinforces that Foundation X, by leveraging knowledge from diverse datasets and utilizing Cyclic and Lock-Release pretraining, boosts performance, even with minimal (head-only) finetuning.

\section{Conclusion}
\label{sec:conclusion}
In this study, we introduce Foundation X, an advanced model for chest X-ray analysis designed to handle classification, localization, and segmentation tasks with a shared backbone. By leveraging the Cyclic and Lock-Release pretraining strategies, Foundation X achieves significant performance improvements across diverse datasets, confirming its capability for combined task learning. Foundation X surpasses baselines across various datasets and tasks while maximizing the utilization of all available annotations. This efficiency reduces annotation costs and enhances the effectiveness of data analysis and processing in medical image analysis. Overall, Foundation X is a robust and versatile solution for advancing medical imaging technology. 
Currently, our work focuses on Chest X-ray images, and we plan to extend it to other imaging modalities as future work.

{\small
\bibliographystyle{ieee_fullname}
\newpage

}
\newpage

\setcounter{page}{1}


\twocolumn[ 
\begin{@twocolumnfalse}
    \begin{center}
    \section*{-- Supplementary Material --}
    \section*{Foundation X: Integrating Classification, Localization, and Segmentation through Lock-Release Pretraining Strategy for Chest X-ray Analysis}
    \end{center}
\end{@twocolumnfalse}
] 

\section{Experiment details}
Here, we discuss the setup of the training process for Foundation X. The backbone is Swin-B, initialized with Ark-6~\cite{ma2023foundation} pretrained weights. For the classification task, linear layers serve as classification heads. For localization, we integrate the DINO localization approach~\cite{zhang2022dino}, modify to handle multiple datasets with one localization encoder and multiple localization decoders. For segmentation, we use UperNet~\cite{xiao2018unifiedUperNet}, modify to include multiple segmentation heads. We pretrain Foundation X on all 11 datasets (see Table 1) using a single A100 GPU, employing the Cyclic and Lock-Release pretraining strategies. We also employ the Student-Teacher learning paradigm, where the teacher model is an exact copy of the student model at the start. The teacher model is updated after each epoch using an exponential moving average (EMA)~\cite{tarvainen2017mean} with a momentum of 0.80. The configuration for Foundation X is detailed in Table~\ref{tab:hyperparameters}.

\begin{table}[h]
\centering
\small
\begin{threeparttable}
\begin{tabular}{p{5.25cm}@{}  p{2.5cm}@{}}
    \midrule
    Backbone & Swin-B$^{\dagger}$~\cite{liu2021swin} \\
    Classification Branch & Linear Layer$^{\ddagger}$ \\
    Localization Branch & DINO$^{\ddagger}$~\cite{zhang2022dino} \\
    Segmentation Branch & UperNet$^{\ddagger}$~\cite{xiao2018unifiedUperNet} \\
    Input Resolution & 224 x 224 \\
    Optimizer & AdamW \\
    Batch Size & 24 \\ 
    Number of Workers & 12 \\ 
    Backbone Learning Rate & 1e-5 \\ 
    Localization Learning Rate & 1e-4 \\ 
    Segmentation Learning Rate & 1e-4 \\ 
    Learning Rate Scheduler & Step-decay \\
    Evaluation Metric for Classification & AUC \\
    Evaluation Metric for Localization & mAP40 \\
    Evaluation Metric for Segmentation & Dice \\
    \bottomrule
\end{tabular}
    \begin{tablenotes}
    \item \footnotesize$^{\dagger}$ Initialized with Ark-6~\cite{ma2023foundation} pretrained weights.
    \item \footnotesize$^{\ddagger}$ Initialized with random weights.
    \end{tablenotes}
    \end{threeparttable}
\caption{Experiment settings for Foundation X.}
\label{tab:hyperparameters}
\end{table}

\begin{table*}[t]
  \footnotesize
  \centering
    \begin{tabular}{ p{1.10cm}@{} P{1cm}@{} P{1.5cm}@{} P{1.5cm}@{} P{1.5cm}@{} P{1.5cm}@{} P{1.5cm}@{} P{1.5cm} p{1.40cm}@{} p{3.5cm}@{} }
    \toprule
             & Epoch \# & Data Size & Backbone & Loc.Enc & Loc.Dec & Seg.Dec & Seg.Head & Mode & Training Task \\  
    \midrule
    \multirow{11}{*}{\fontsize{9}{14}\selectfont \textbf{Cycle 1}}  & 1 & Half & F & F & T & - & - & Lock    & Localization of Heart \\
             & 2 & Full & T & T & T & - & - & Release & Localization of Heart \\
             & 3 & Half & F & F & T & - & - & Lock    & Localization of Left Lung \\
             & 4 & Full & T & T & T & - & - & Release & Localization of Left Lung \\
             & 5 & Half & F & F & T & - & - & Lock    & Localization of Right Lung \\
             & 6 & Full & T & T & T & - & - & Release & Localization of Right Lung \\

             & 7  & Half & F & - & - & F & T & Lock    & Segmentation of Heart \\
             & 8  & Full & T & - & - & T & T & Release & Segmentation of Heart \\
             & 9  & Half & F & - & - & F & T & Lock    & Segmentation of Left Lung \\
             & 10 & Full & T & - & - & T & T & Release & Segmentation of Left Lung \\
             & 11 & Half & F & - & - & F & T & Lock    & Segmentation of Right Lung \\
             & 12 & Full & T & - & - & T & T & Release & Segmentation of Right Lung \\
    \bottomrule
  \end{tabular}
  \caption{Demonstrating the Lock-Release pretraining strategy for organ localization and segmentation using the VinDr-CXR dataset. The model completes a single cycle when it goes through all tasks once (from epoch \#1 to \#12). 'F' denotes a frozen component, and 'T' denotes a trainable component. In Lock mode, the model is trained using half of the dataset, while in Release mode, it is trained using the full dataset. After each epoch in Release mode, the model is tested on the localization and segmentation of the heart, left lung, and right lung.}
  \label{tab:trainingstrategyOrganLoc}
\end{table*}

\section{Model Parameters}
The Foundation X model consists of several key components, contributing to a total of approximately 173 million parameters (See Table~\ref{tab:parameter_distribution}). The backbone, responsible for feature extraction, accounts for 86.8 million parameters. The localization encoder adds 7.7 million parameters, while the localization decoders total 57.6 million, with each decoder contributing 9.6 million. The segmentation decoder comprises 20.9 million parameters.
Although adding a dedicated localization decoder for each dataset increases the model size, only the relevant decoder is active during training, with the others, along with the classification and segmentation heads, remaining frozen. This approach keeps the computational load manageable and ensures efficient GPU utilization.
\begin{table}[h]
\centering
\begin{tabular}{l|r}
\hline
\textbf{Component} & \textbf{Parameters} \\ \hline
Backbone           & 86,751,673 [+]         \\  
Classification Heads   & 70,725 [+]         \\  
Localization Encoder  & 7,693,056 [+]         \\ 
Localization Decoders  & 57,653,868 [+] \\
\quad \textit{Each Localization Decoder} & \textit{9,608,978} \\ 
Segmentation Decoder   & 20,894,464 [+]      \\ 
Segmentation Heads   & 20,754 [+]      \\ \hline
\textbf{Total}     & \textbf{173,084,540} \\ \hline
\end{tabular}
\caption{Parameter distribution across the key components of the Foundation X model, trained on 11 datasets and 20 tasks.}
\label{tab:parameter_distribution}
\vspace{-1em}
\end{table}

\section{Lock-Release pretraining strategy}
Foundation X effectively handles classification, localization, and segmentation tasks. The model leverages the Student-Teacher learning paradigm along with Cyclic and Lock-Release pretraining strategies, ensuring it retains general knowledge for all tasks while avoiding overfitting to any single task. In Table~\ref{tab:trainingstrategyOrganLoc}, we illustrate the Lock-Release pretraining strategy using the VinDr-CXR organ dataset for organ localization and segmentation. For this demonstration, we treat localization and segmentation of the heart, left lung, and right lung as separate tasks.

\begin{algorithm*}
\DontPrintSemicolon
\KwData{Datasets: $\mathcal{D}=\{D_1, D_2, ..., D_n\}$; Sample: image-label pair $(x,y) \in \mathcal{D}_i$}
\KwFunctions{Data augmentation: $\varepsilon$; Dataset/task-specific losses: $\{\mathcal{L}{D1}(\cdot,\cdot), \mathcal{L}{D2}(\cdot,\cdot), ..., \mathcal{L}{Dn}(\cdot,\cdot)\}$; Consistency loss: $\{\mathcal{L}{const}(\cdot,\cdot)\}$; Loss update by AdamW optimizer: $Update_{adamw}(\cdot,\cdot)$ }
\KwTrainables{Student's encoder, localization encoder, segmentation decoder: $e_s$, $LocEnc_s$, $SegDec_s$;  Classification heads $\mathcal{C} = \{C_1, C_2, ..., C_n\}$; Localization decoders $\mathcal{L} = \{L_1, L_2, ..., L_n\}$; Segmentation heads $\mathcal{S} = \{S_1, S_2, ..., S_n\}$; }
\KwStopgrad{Teacher's encoder, localization encoder, segmentation decoder: $e_t$, $LocEnc_t$, $SegDec_t$; }
\KwHyperparameters{Momentum: $\lambda$}

 $\{e_t,LocEnc_t,SegDec_t\} \leftarrow \{e_s,LocEnc_s,SegDec_s\}$ \tcp*{initialize teacher with student's parameters}

\For{$D_i$ \textbf{in} ${D_1, D_2, ..., D_n}$}   
{ 
    \tcc{train student for one epoch}
    \For{$(x,y)$ \textbf{in} $D_i$} 
    {
        $x' = \varepsilon(x)$\;
        \If{$D_i$ has Classification Annotation}
        {   
            \For{$j \gets 1$ \KwTo 2}
            {
                \If{$j$ = 1}
                    { $Freeze$ $\{e_s\}$ \hspace{1cm}\tcp{Lock mode on, using a random half of the dataset} } 
                \Else{ $Unfreeze$ $\{e_s\}$ \hspace{1cm}\tcp{Release mode on, using full dataset} }
                $emb_t,  emb_s = e_t(x'),  e_s(x')$\;
                $pred = C_i(emb_s)$\;
                $Loss = \mathcal{L}{Di}(pred, y) + \mathcal{L}{const1}(emb_t,emb_s)$ \;
                $Update(\{e_s, p_s, C_i\}, Loss) $\;      
            }
        }    
        \If{$D_i$ has Localization Annotation}
        {   
            \For{$j \gets 1$ \KwTo 2}
            {
                \If{$j$ = 1}
                    { $Freeze$ $\{e_s,LocEnc_s\}$ \hspace{1cm}\tcp{Lock mode on, using a random half of the dataset} }
                \Else{ $Unfreeze$ $\{e_s,LocEnc_s\}$ \hspace{1cm}\tcp{Release mode on, using full dataset} }
            $emb_t,  emb_s = e_t(x'),  e_s(x')$\;
            $embLocEnc_s,  embLocEnc_t = LocEnc_s(emb_s),  LocEnc_t(emb_t)$\;
            $pred = L_i(embLocEnc_s)$\;
            $Loss = \mathcal{L}{Di}(pred, y) + \mathcal{L}{const}(emb_t,emb_s) + \mathcal{L}{const}(embLocEnc_t,embLocEnc_s)$ \;
            $Update(\{e_s, LocEnc_s, L_i\}, Loss) $\;
            }
        } 
        \If{$D_i$ has Segmentation Annotation}
        {
            \For{$j \gets 1$ \KwTo 2}
            {
                \If{$j$ = 1}
                    { $Freeze$ $\{e_s,SegDec_s\}$ \hspace{1cm}\tcp{Lock mode on, using a random half of the dataset} }
                \Else{ $Unfreeze$ $\{e_s,SegDec_s\}$ \hspace{1cm}\tcp{Release mode on, using full dataset} }
            $emb_t,  emb_s = e_t(x'),  e_s(x')$\;
            $embSegDec_s,  embSegDec_t = SegDec_s(emb_s),  SegDec_t(emb_t)$\;
            $pred = S_i(embSegDec_s)$\;
            $Loss = \mathcal{L}{Di}(pred, y) + \mathcal{L}{const}(emb_t,emb_s) + \mathcal{L}{const}(embSegDec_t,embSegDec_s)$ \;
            $Update(\{e_s, SegDec_s, S_i\}, Loss) $\;
            }
        }         
    }
    \tcc{Update teacher by student's parameters via epoch-wise EMA}
    $\{e_t, LocEnc_t, SegDec_t\} \leftarrow \lambda  \{e_t, LocEnc_t, SegDec_t\} + (1-\lambda) \{e_s, LocEnc_s, SegDec_s\}$

}
\caption{A round of Foundation X's Cyclic Lock-Release pretraining}\label{alg:foundationX}
\end{algorithm*}

\section{Cross-Dataset and Cross-Task learning analysis}
\begin{figure*}[h]
    \centering
    \includegraphics[width=0.95\textwidth]{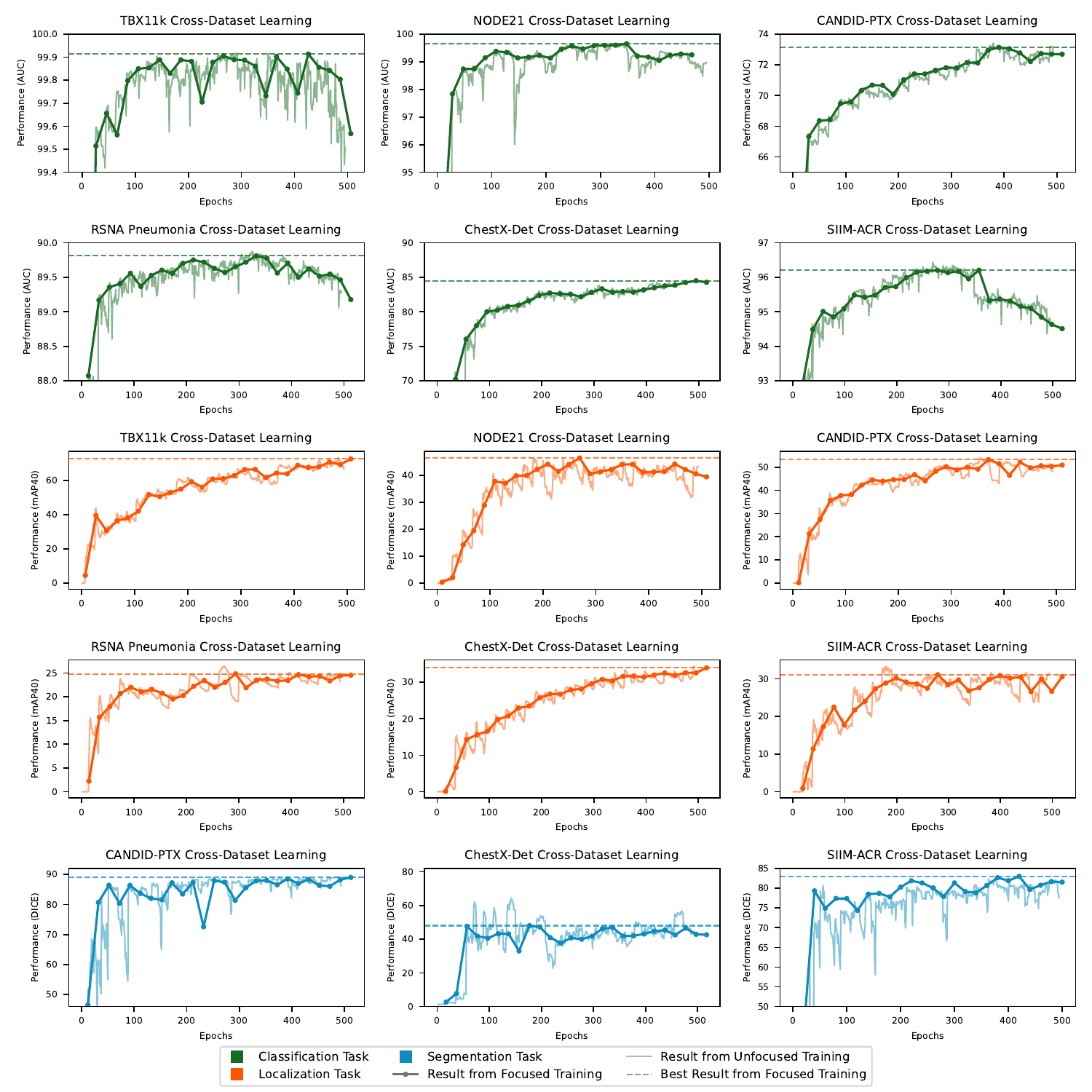}
\caption{Cross-Dataset \& Cross-Task Learning Analysis. The figure demonstrates the performance trends of Foundation X across multiple datasets for both focused and unfocused training scenarios. Focused training refers to scenarios where the model is explicitly trained on the specific dataset being evaluated, while unfocused training refers to scenarios where the model is trained on other datasets and not directly on the dataset being evaluated. The green, orange, and blue lines represent the classification, localization, and segmentation tasks, respectively. Dark-colored lines indicate the testing results during focused training, where the model is explicitly trained on the specific dataset. Light-colored lines show the testing results during unfocused training, where the model is trained on other datasets but tested on the specific dataset. Dashed lines represent the best testing results achieved from focused training for each specific dataset. The results indicate that, during unfocused training, initial performance dips are common as the model is not explicitly trained on the specific dataset. However, performance improves over time, demonstrating the model’s ability to generalize effectively and retain knowledge due to the Cyclic and Lock-Release pretraining strategies. In all cases, the unfocused training results do not drift away from the task, highlighting the model’s efficient generalization and knowledge retention. Additionally, in some instances, unfocused training achieves even better performance than focused training, showcasing the advantages of cross-task and cross-dataset learning in enhancing the overall capabilities of Foundation X.}
    \label{X_CrossDataset_CrossTask2_FULL}
\end{figure*}

The full figure illustrating the Cross-Dataset and Cross-Task learning analysis for all six datasets is included in this supplementary material (see Figure~\ref{X_CrossDataset_CrossTask2_FULL}). This figure highlights the performance trends of Foundation X across various datasets under both focused and unfocused training scenarios, showcasing its ability to generalize and retain knowledge effectively through the Cyclic and Lock-Release pretraining strategies. We include plots for these six datasets because they contain multiple tasks, including classification, localization, and segmentation.

\section{Ablation study}
The ablation studies demonstrate the effectiveness of the Student-Teacher learning paradigm, Cyclic and Lock-Release pretraining strategies across various tasks. Foundation X-L (see Table~\ref{tab:pretrainingXloc}), trained on six disease localization tasks, generally outperforms the baseline model Swin-B + DINO. Similarly, Foundation X-S (see Table~\ref{tab:pretrainingXseg}), trained on three disease segmentation datasets, consistently surpasses the baseline model Swin-B + UperNet. Additionally, Foundation X-CL (see Table~\ref{tab:pretrainingXclsloc}), which handles both classification and localization tasks, and Foundation X-LS (see Table~\ref{tab:pretrainingXlocseg}), which integrates localization and segmentation tasks, both show superior performance compared to their respective baseline methods in most cases.

The ablation study presented in Table~\ref{tab:ablation_study_LR_ST} highlights the impact of incorporating the Lock-Release pretraining strategy and the Student-Teacher learning paradigm on the performance of the Foundation X model. The results demonstrate that when both components are enabled, the model achieves the best performance across all evaluated VinDr-CXR organs (Heart, Left Lung, and Right Lung) localization. Specifically, the combination of Lock-Release and Student-Teacher results in the highest mAP, with scores of 88.39\% for Heart, 95.78\% for Left Lung, and 96.78\% for Right Lung. 
These findings suggest that each component complements the other, with the Lock-Release strategy preventing task-specific overfitting and the Student-Teacher paradigm ensuring stable learning by reducing drastic model shifts. Together, these strategies create a synergistic effect that enhances the model’s generalization and overall performance, outperforming configurations where one or both components are disabled. This highlights the importance of integrating both the Lock-Release strategy and the Student-Teacher paradigm to maximize the effectiveness of our approach.

\begin{table}[h]
    \centering
    \begin{tabular}{P{1cm}@{} P{2cm}@{} P{1cm}@{} P{1.7cm}@{} P{1.7cm}@{}}
        \toprule
        Lock-Release & Student-Teacher & Heart & Left Lung & Right Lung \\
        \midrule
        \xmark     & \xmark     & 85.45 & 93.63 & 94.47 \\
        \xmark     & \checkmark & 86.41 & 94.39 & 94.95 \\
        \checkmark & \xmark     & 87.50 & 95.37 & 96.44 \\
        \checkmark & \checkmark & \textbf{88.39} & \textbf{95.78} & \textbf{96.78} \\
        \bottomrule
    \end{tabular}
    \caption{{Ablation study is conducted on the VinDr-CXR organ localization dataset. We evaluate the model with and without the Lock-Release pretraining strategy, as well as with and without the Student-Teacher model. The results demonstrate that the Foundation X model achieves comparatively better performance when both the Lock-Release pretraining strategy and the Student-Teacher learning paradigm are employed.}}
    \label{tab:ablation_study_LR_ST}
    \vspace{-1em}
\end{table}

\begin{table*}[h]
  \small
  \centering
  \begin{threeparttable}
  \begin{tabular}{ p{3cm}@{} P{3cm}@{} P{4cm}@{} }
    \toprule
    Dataset & Baseline Loc.$^{\dagger}$ & Foundation X-L \\
            & \scriptsize[mAP40\%] & \scriptsize[mAP40\%] \\         
    \midrule
    TBX11K         & \textbf{78.10} & 77.77 {\footnotesize\NIred{$\downarrow0.33$}}\\
    NODE21         & 37.50          & \textbf{42.79 }{\footnotesize\NIgreen{$\uparrow5.29$}} \\
    CANDID-PTX     & 50.90          & \textbf{53.75 }{\footnotesize\NIgreen{$\uparrow2.85$}} \\
    RSNA Pneumonia & 21.70          & \textbf{29.37 }{\footnotesize\NIgreen{$\uparrow7.67$}} \\
    ChestX-Det     & 38.00          & \textbf{40.13 }{\footnotesize\NIgreen{$\uparrow2.13$}} \\
    SIIM-ACR       & 28.00          & \textbf{36.20 }{\footnotesize\NIgreen{$\uparrow8.20$}} \\
    \bottomrule
  \end{tabular}
    \begin{tablenotes}
    \item \footnotesize$^{\dagger}$ Swin-B version of DINO where the backbone is initialized with Ark-6 pretrained weights.
    \end{tablenotes}
    \end{threeparttable}
  \caption{We train Foundation X-L on six disease localization tasks utilizing Cyclic and Lock-Release pretraining strategies and compare its performance with the baseline model, DINO~\cite{zhang2022dino}. In most cases, Foundation X-L outperforms the baseline across the datasets during pretraining.}
  \label{tab:pretrainingXloc}
\end{table*}

\begin{table*}[h]
  \small
  \centering
  \begin{threeparttable}
  \begin{tabular}{ p{3cm}@{} P{2cm}@{} P{3cm}@{} }
    \toprule
    Dataset & Baseline Seg.$^{\dagger}$ & Foundation X-S \\
            & \scriptsize[Dice\%] & \scriptsize[Dice\%] \\   
    \midrule
    CANDID-PTX & 86.36 & \textbf{89.58 }{\footnotesize\NIgreen{$\uparrow3.23$}} \\
    ChestX-Det & 79.33 & \textbf{83.46 }{\footnotesize\NIgreen{$\uparrow4.13$}} \\
    SIIM-ACR   & 81.92 & \textbf{83.83 }{\footnotesize\NIgreen{$\uparrow1.91$}} \\
    \bottomrule
  \end{tabular}
    \begin{tablenotes}
    \item \footnotesize$^{\dagger}$ Swin-B version of UperNet where the backbone is initialized with Ark-6 pretrained weights.
    \end{tablenotes}
    \end{threeparttable}
  \caption{We train Foundation X-S on three disease segmentation datasets using the Cyclic and Lock-Release pretraining strategies and compare its performance with the baseline model, UperNet~\cite{xiao2018unifiedUperNet}. In all cases, Foundation X-S outperforms the baseline across the datasets during pretraining.}
  \label{tab:pretrainingXseg}
\end{table*}

\begin{table*}[h]
  \small
  \centering
  \begin{tabular}{ p{3cm}@{} P{2.25cm}@{} P{2.25cm}@{} P{2cm}@{} P{3cm}@{} P{3cm}@{} }
    \toprule
    Dataset & Baseline Cls. & Baseline Loc. & \multicolumn{2}{c}{Foundation X-CL} \\
            & \scriptsize[AUC\%] & \scriptsize[mAP40\%] & \scriptsize[AUC\%] & \scriptsize[mAP40\%] \\   
    \midrule
    TBX11K & 99.89\footnotesize$\pm0.06$ & \textbf{78.10} & \textbf{99.96 }{\footnotesize\NIgreen{$\uparrow0.07$}} & {72.56 }{\footnotesize\NIred{$\downarrow5.54$}} \\
    NODE21 & 99.35\footnotesize$\pm0.45$ & 37.50 & \textbf{99.68 }{\footnotesize\NIgreen{$\uparrow0.33$}} & \textbf{47.54 }{\footnotesize\NIgreen{$\uparrow10.04$}} \\
    CANDID-PTX & 72.61\footnotesize$\pm0.57$ & 50.90 & \textbf{74.00 }{\footnotesize\NIgreen{$\uparrow1.39$}} & \textbf{51.61 }{\footnotesize\NIgreen{$\uparrow0.71$}} \\
    RSNA Pneumonia & 88.87\footnotesize$\pm0.21$ & 21.70 & \textbf{96.57 }{\footnotesize\NIgreen{$\uparrow7.70$}} & \textbf{26.08 }{\footnotesize\NIgreen{$\uparrow4.38$}} \\
    ChestX-Det & \textbf{88.17}\footnotesize$\pm0.33$ & \textbf{38.00} & {81.82 } {\footnotesize\NIred{$\downarrow6.35$}} & {37.03 }{\footnotesize\NIred{$\downarrow0.97$}} \\
    SIIM-ACR & 95.01\footnotesize$\pm0.16$ & 28.00 & \textbf{95.19 }{\footnotesize\NIgreen{$\uparrow0.18$}} & \textbf{34.98 }{\footnotesize\NIgreen{$\uparrow6.98$}} \\
    \bottomrule
  \end{tabular}
  \caption{We train Foundation X-CL on six disease datasets, each containing both classification and localization annotations, using our Cyclic and Lock-Release pretraining strategies. The table demonstrates that, in most cases, Foundation X-CL outperforms the baseline methods during pretraining.}
  \label{tab:pretrainingXclsloc}
\end{table*}

\begin{table*}[h]
  \small
  \centering
  \begin{tabular}{ p{3cm}@{} P{2.5cm}@{} P{2cm}@{} P{2cm}@{} P{2.5cm}@{} P{2.5cm}@{} }
    \toprule
    Dataset & Baseline Loc. & Baseline Seg. & \multicolumn{2}{c}{Foundation X-LS} \\
            & \scriptsize[mAP40\%] & \scriptsize[Dice\%] & \scriptsize[mAP40\%] & \scriptsize[Dice\%] \\   
    \midrule
    TBX11K         & \textbf{78.10} & - & {73.03 }{\footnotesize\NIred{$\downarrow5.07$}} & - \\
    NODE21         & 37.50 & - & \textbf{46.09 }{\footnotesize\NIgreen{$\uparrow8.59$}} & - \\
    CANDID-PTX     & 50.90 & 86.36 & \textbf{53.01 } {\footnotesize\NIgreen{$\uparrow2.11$}} & \textbf{89.47 }{\footnotesize\NIgreen{$\uparrow3.11$}} \\
    RSNA Pneumonia & 21.70 & - & \textbf{27.80 }{\footnotesize\NIgreen{$\uparrow6.10$}} & - \\
    ChestX-Det     & 38.00 & \textbf{79.33} & \textbf{39.22 }{\footnotesize\NIgreen{$\uparrow1.22$}} & {70.90 }{\footnotesize\NIred{$\downarrow8.43$}} \\
    SIIM-ACR       & 28.00 & 81.92 & \textbf{36.63 }{\footnotesize\NIgreen{$\uparrow8.63$}} & \textbf{84.25 }{\footnotesize\NIgreen{$\uparrow2.33$}} \\
    \bottomrule
  \end{tabular}
  \caption{We train Foundation X-LS on six disease localization and three disease segmentation datasets, using our Cyclic and Lock-Release pretraining strategies. The table demonstrates that, in most cases, Foundation X-LS outperforms the baseline methods during pretraining.}
  \label{tab:pretrainingXlocseg}
\end{table*}

\begin{table*}[h]

\small
\centering
\begin{tabular}{ P{1.7cm}@{} |p{3.5cm}@{} P{1cm}@{} P{2cm}@{} P{1.5cm}@{} P{1.5cm}@{} P{5cm}@{} }
\hline
Task & Dataset & Official Split & Train Split & Val Split & Test Split & Expert Labels\\
\hline
\multirow{22}{*}{\makecell[c]{CLS}} & CheXpert~\cite{irvin2019chexpert} & \checkmark & 223415 & 234 & - & No finding, Enlarged Cardiomediastinum, Cardiomegaly, Lung Opacity, Lung Lesion, Edema, Consolidation, Pneumonia, Atelectasis, Pneumothorax, Pleural Effusion, Pleural Other, Fracture, Support Devices \\\cline{2-7}
  & NIH ChestX-ray14~\cite{wang2017chestx} & \checkmark  & 75312 & 11212 & 25596 & Atelectasis, Cardiomegaly, Effusion, Infiltration, Mass, Nodule, Pneumonia, Pneumothorax, Consolidation, Edema, Emphysema, Fibrosis, Pleural Thickening, Hernia \\\cline{2-7}
  & VinDr-CXR~\cite{nguyen2022vindr} & \checkmark  & 15000 &  -  & 3000 & PE, Lung Tumor, Pneumonia, Tuberculosis, Other diseases, No finding \\\cline{2-7}
  & NIH Shenzhen CXR~\cite{jaeger2014two} & \ding{55}  & 463 & 65 & 134 & Tuberculosis \\\cline{2-7}
  & MIMIC-II~\cite{johnson2019mimic} & \checkmark  & 368878 & 2991 & 5159 & No finding, Enlarged Cardiomediastinum, Cardiomegaly, Lung Opacity, Lung Lesion, Edema, Consolidation, Pneumonia, Atelectasis, Pneumothorax, Pleural Effusion, Pleural Other, Fracture, Support Devices \\ \hline 
  \multirow{5}{*}{\makecell[c]{CLS, LOC}}& TBX11k~\cite{liu2020rethinking} & \checkmark  & 6600 & 1800 & - & Tuberculosis\\\cline{2-7}
  & NODE21~\cite{sogancioglu2024nodule} & \ding{55} & 4178 & - & 1046 & Nodule \\\cline{2-7}
  & RSNA Pneumonia~\cite{radiological2018rsna} & \checkmark  & 21295 & 2680 & 2709 & CLS: No lung opacity/Not normal, Normal, Lung Opacity; LOC: Pneumonia \\ \hline 
 \multirow{7}{*}{\makecell[c]{CLS, LOC, \\SEG}} & CANDID-PTX~\cite{feng2022automated} & \checkmark & 13748 & 1964 & 3928 & Pneumothorax \\\cline{2-7}
  & ChestX-Det~\cite{lian2021structure} & \checkmark  & 3025 & - & 553 & Atelectasis, Calcification, Cardiomegaly, Consolidation, Diffuse Nodule, Effusion, Emphysema, Fibrosis, Fracture, Mass, Nodule, Pleural Thickening, Pneumothorax \\ \cline{2-7}
  & SIIM-ACR ~\cite{dramsch2019siim} & \ding{55} & 9607 & 1068 & 1372 & Pneumothorax \\ 
\hline 
& \textbf{Total CLS images} & & \textbf{741521} & \textbf{22014} &  \textbf{43997} & \\  
& \textbf{Total LOC images} & & \textbf{58453} & \textbf{7512} &  \textbf{9608} & \\  
& \textbf{Total SEG images} & & \textbf{26380} & \textbf{3032} &  \textbf{5853} & \\  
\hline \hline

\multirow{6}{*}{\makecell[c]{LOC FT}} & VinDr-CXR~\cite{nguyen2022vindr} & \checkmark & 15000 & - & 3000 & Aortic enlargement, Atelectasis, Calcification, Cardiomegaly, Consolidation, ILD, Infiltration, Lung Opacity, Nodule/Mass, Other lesion, Pleural effusion, Pleural thickening, Pneumothorax, Pulmonary fibrosis \\ 
\hline
\multirow{4}{*}{\makecell[c]{SEG FT}} & CheXmask VinDr-CXR~\cite{gaggion2024chexmask} & \checkmark & 15000 & - & 3000 & Heart, Left Lung, Right Lung \\ \cline{2-7}
    & VinDr-RibCXR~\cite{nguyen2021vindrRIB} & \checkmark & 196 & - & 49 & 20 Ribs \\ \cline{2-7}
    & NIH Montgomery~\cite{jaeger2014two} & \ding{55} & 92 & 15 & 31 & Lung \\ \cline{2-7}
    & JSRT~\cite{van2006segmentation} & \ding{55} & 173 & 25 & 49 & Heart, Lung, Clavicle \\
\hline
\end{tabular}

\vspace{0.5cm}
\caption{Foundation X was pretrained on the above 11 classification datasets, 6 localization datasets, and 3 segmentation datasets. CLS stands for classification task, LOC stands for localization task, SEG stands for segmentation task. "CLS, LOC" denotes the datasets used for classification and localization tasks. "CLS, LOC, SEG" denotes the datasets used for classification, localization, and segmentation tasks. "LOC FT" and "SEG FT" denotes the datasets used only during the finetuning of the localization and segmentation task, respectively.}
\end{table*}

\section{Acknowledgements}
This research was partially supported by ASU and Mayo Clinic through a Seed Grant and an Innovation Grant, as well as by the NIH under Award Number R01HL128785. The authors are solely responsible for the content, which does not necessarily reflect the official views of the NIH. This work also utilized GPUs provided by ASU Research Computing, Bridges-2 at the Pittsburgh Supercomputing Center (allocated under BCS190015), and Anvil at Purdue University (allocated under MED220025). These resources are supported by the Advanced Cyberinfrastructure Coordination Ecosystem: Services \& Support (ACCESS) program, funded by the National Science Foundation under grants \#2138259, \#2138286, \#2138307, \#2137603, and \#2138296. We also extend our gratitude to Anirudh Kaniyar Narayana Iyengar for his contributions to collecting localization data, preparing bounding boxes in COCO format, and developing some of the dataloaders. Finally, the content of this paper is covered by patents pending.
\end{document}